# Reading Between the Lines: Classifying Resume Seniority with Large Language Models

Matan Cohen
School of Computer Science,
Faculty of Sciences
Holon Institute of Technology

Shira Shani
School of Computer Science,
Faculty of Sciences
Holon Institute of Technology

Eden Menahem
School of Computer Science,
Faculty of Sciences
Holon Institute of Technology

Yehudit Aperstein
Intelligent Systems,
Academic College of Engineering
Tel Aviv Israel

Alexander Apartsin
School of Computer Science,
Faculty of Sciences
Holon Institute of Technology

*Abstract*—Accurately assessing candidate seniority from resumes is a critical yet challenging task, complicated by the prevalence of overstated experience and ambiguous self-presentation. In this study, we investigate the effectiveness of large language models (LLMs), including fine-tuned BERT architectures, for automating seniority classification in resumes. To rigorously evaluate model performance, we introduce a hybrid dataset comprising both real-world resumes and synthetically generated hard examples designed to simulate exaggerated qualifications and understated seniority. Using the dataset, we evaluate the performance of Large Language Models in detecting subtle linguistic cues associated with seniority inflation and implicit expertise. Our findings highlight promising directions for enhancing AI-driven candidate evaluation systems and mitigating bias introduced by self-promotional language. The dataset is available for the research community at https://bit.ly/4mcTovt

## I. Introduction

The classification of professional seniority based on resume content is a fundamental task in talent acquisition and workforce analytics. Accurate identification of a candidate's seniority level has a direct impact on hiring decisions, compensation benchmarks, and career development planning. However, this task remains challenging due to the inherently subjective and often strategic nature of self-presentation in resumes. Candidates frequently employ persuasive language, highlight selective experiences, and utilize industry-specific jargon to portray themselves as possessing higher levels of expertise. In contrast, others may modestly understate their capabilities despite possessing substantial qualifications. These practices introduce significant ambiguity, making automated analysis of seniority a non-trivial problem.

With the advent of large language models (LLMs), there is growing interest in leveraging their advanced natural language understanding capabilities for resume analysis and candidate evaluation. Models such as BERT and its variants have demonstrated strong performance in various text classification and semantic understanding tasks. However, their ability to detect nuanced patterns of self-promotion, exaggeration, and understated experience in resumes remains underexplored. In high-stakes environments such as recruitment, where fairness, accuracy, and bias mitigation are paramount, it is essential to rigorously assess the robustness of these models against complex linguistic manipulations.

To address this gap, we propose a comprehensive evaluation framework for automated seniority classification utilizing large language models (LLMs). We introduce a hybrid dataset that combines authentic resumes with a carefully constructed set of synthetic "hard examples," explicitly designed to simulate cases of overstated experience and understated seniority. This unique dataset challenges models to move beyond superficial text patterns and instead identify subtle linguistic cues indicative of actual professional standing.

The main contributions of this work are as follows:

• Hybrid benchmark dataset: We introduce a novel dataset for seniority classification, combining real-world resumes with synthetically generated hard examples that explicitly model overstated experience and understated seniority scenarios.

• Comparison of zero-shot GPT-based and fine-tuned BERT-based Large Language Models for resume seniority classification: We evaluate multiple state-of-the-art models, including zero-shot classification with GPT 4.0 and two fine-tuned BERT variants, and compare their performance to a traditional logistic regression model based on extracted TF-IDF features.

## II. Literature Review

LLMs have been applied to automate and enhance traditional resume screening processes. For example, Lo et al. (2025) propose a multi-agent framework using LLMs (with retrieval-augmented generation, RAG) to extract structured information, score candidates, and generate summaries from resumes. Their system divides tasks into specialized agents (extractor, evaluator, summarizer, etc.), enabling role-specific scoring and the incorporation of external knowledge (e.g., industry criteria) to refine assessments. Similarly, Gan et al. (2024) introduce an LLM-agent framework that summarizes and grades large resume datasets. In simulations using real resumes, they report that their system is 11 times faster than manual screening and achieves high accuracy (87.7% F1 in resume sentence classification) by fine-tuning large language models (LLMs). Such studies highlight LLMs' potential to automate labor-intensive steps in talent acquisition, including resume parsing, skills matching, and ranking, while providing explainable feedback.

Several studies compare the performance of LLMs to that of human screening. Vaishampayan et al. (2025) conducted an observational study of 736 real applicants, comparing zero-shot GPT-4 ratings to those of human recruiters across various criteria, such as experience and skills. They find that LLM scores only weakly correlate with human judgments, implying LLMs are not interchangeable with people. However, careful prompt design (e.g., chain-of-thought) improves alignment. Overall, these efforts demonstrate LLMs can parse, classify, and score resumes automatically, often with efficiency gains, but may require calibration to align with human decision standards.

LLMs can transform unstructured resumes into structured profiles, extract skills, and generate candidate summaries. Frameworks like those of Lo et al. (2025) and Gan et al. (2024) utilize multi-agent large language model (LLM) architectures to emulate stages of hiring, including parsing, scoring, and summarizing, thereby yielding scalable and explainable screening systems. Empirical studies show LLM-driven screening is much faster than manual review (e.g., 11× speed-up) and can achieve high classification scores after fine-tuning. However, LLM outputs must be carefully validated against human criteria (e.g., Vaishampayan et al., 2025).

LLMs have also been applied to classify resume content and recognize candidate expertise. Heakl et al. (2024) curate a large dataset of 13,389 resumes and evaluate transformer models (e.g., BERT and "Gemma1.1") for categorizing resumes. Their best LLM-based classifier achieves ~92% top-1 accuracy (and 97.5% top-5), far surpassing earlier shallow methods. This "ResuméAtlas" study highlights that high-quality, diverse training data, combined with modern large language models (LLMs), significantly enhances the categorization of resumes. In a related vein, Skill-LLM (Herandi et al., 2024) fine-tunes a transformer specifically to extract *skill entities* from job-related text. Their specialized Skill-LLM outperforms prior state-of-the-art on benchmark skill-extraction tasks, showing that LLMs (with domain-adapted training) can enhance precision in parsing candidate expertise.

Beyond raw text classification, LLMs can aid in profiling expertise and identifying areas of specialization. For example, prompting a powerful LLM like GPT-4 to analyze a resume can reveal implicit skills or qualifications that are not explicitly listed. LLMs' broad knowledge enables them to infer transferable skills (e.g., recognizing coding skills implied by project descriptions). In summary, recent research has found that transformer-based language models (LLMs) excel at categorizing resumes and identifying professional skills. They benefit from large, labelled resume corpora (e.g., Heakl et al. 2024) and fine-tuning or prompt engineering (Herandi et al., 2024) to capture domain-specific expertise.

Job applicants often engage in strategic self-presentation, using positive framing or embellishments in resumes. NLP research has begun to address **linguistic deception** and bias in such contexts. While domain-specific studies on resume lies are scarce, related work on textual deception provides insight. Tomas et al. (2022) review computational approaches to deception detection, noting familiar cues: deceptive texts often contain more negations, fewer self-references, and less detailed content than truthful ones. Transformer-based models (e.g., BERT) can modestly improve detection; for example, Ilias et al. (2022) combine BERT and LSTM architectures to detect deception in statements, observing a nearly 2% accuracy gain over simpler models. They further use explainability (LIME) to identify that deceptive language differs significantly in LIWC categories (e.g., affect words, pronouns) from truthful language.

Emerging work explores the role of LLMs in both enabling and detecting deception. Armstrong et al. (2024) demonstrate that GPT-3.5 encodes social stereotypes. In auditing resumes with varied names, GPT assigned higher ratings to resumes with White male names and generated biased content (e.g., Asian/Hispanic candidates with "immigrant" markers). This suggests LLMs can unwittingly reinforce self-presentational biases. Conversely, detection systems could leverage LLMs to flag manipulative writing, although this remains a nascent area. In interviews, reporters note that some NLP tools attempt to identify exaggerations or inconsistencies in cover letters, but robust methods are underdeveloped.

Using large language models (LLMs) for hiring raises critical concerns about bias, fairness, and robustness. Several studies have audited LLM-based hiring tools for demographic biases. Armstrong et al. (2024) find that GPT-3.5 exhibits racial and gender biases when scoring resumes. For example, it was less likely to "interview" candidates with non-White names and generated more stereotypical resumes for specific groups. Wilson and Caliskan (2024) simulate resume screening with semantic embedding models and report alarming disparities: their system favored White-associated names in 85.1% of cases and rarely favored female-associated names (11.1%). Black male candidates were especially disadvantaged, echoing real-world hiring biases. In contrast, Vaishampayan et al. (2025) found that GPT-4's resume ratings did not exhibit more substantial racial or gender group differences than human raters; they suggest that careful chain-of-thought prompting can further reduce bias.

Researchers are also proposing fairness-aware frameworks. Haryan et al. (2024) introduce "FairHire," an automated screening system designed to enforce equitable treatment, though details remain limited. A broad survey by Mujtaba and Mahapatra (2024) outlines metrics and mitigation strategies across the recruitment pipeline, emphasizing preprocessing (e.g., anonymizing protected attributes), in-processing adjustments, and post-hoc audits. For example, removing names and sensitive information can improve fairness, and regular audits, as done by Armstrong et al. (2024), help detect embedded biases. Robustness also involves handling adversarial inputs; while not yet deeply studied in hiring, some authors caution that LLMs could be tricked by maliciously phrased resumes, underscoring the need for robustness testing.

III. METHODOLOGY

A. Dataset construction

The synthetic dataset used in this study was specifically designed to assess the ability of large language models to classify professional seniority and handle various forms of linguistic manipulation. Using the Mistral-7B model, we generated synthetic resumes that simulate different self-presentation strategies by creating three parallel versions of each base resume scenario: one reflecting normal seniority, one intentionally understating seniority, and one overstating it. This parallel structure ensures that models are tested on their capacity to distinguish between subtle linguistic variations while the core factual content remains consistent across all versions.

The process began by defining a representative set of job roles across multiple industries, such as Software Engineer, Project Manager, and Data Scientist. For each role, we compiled a list of relevant skills and typical career experiences based on labor market data and publicly available job descriptions. GPT-4o was then prompted to generate a baseline resume accurately reflecting the expected qualifications for the role.

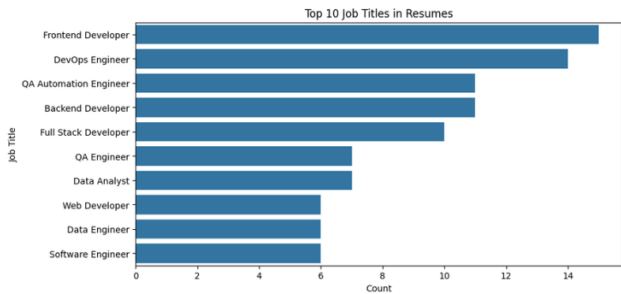

*Figure 1*: Top 10 job titles of the resumes in the generated dataset

| Version | Prompt |
|---|---|
| Normal Seniority | "Generate a professional resume for a Software Engineer named Alex Johnson with 5 years of experience in the tech industry. Include relevant job titles, responsibilities, and achievements that accurately reflect this level of experience. Use clear and factual language without exaggeration. Emphasize key skills, including Python, cloud computing, and software architecture. Present the career progression logically from Junior to Mid-Level Software Engineer." |
| Understated Seniority | "Generate a professional resume for a Software Engineer named Alex Johnson who has 5 years of experience but prefers to present their qualifications modestly. Use humble language and downplay achievements. Avoid using senior-sounding titles or strong promotional phrases. Focus on collaborative contributions rather than leadership roles. Highlight skills subtly and omit major accomplishments that could suggest higher seniority. Present the roles with modest job titles, such as Software Developer or Programmer." |
| Overstated Seniority | "Generate a professional resume for a Software Engineer named Alex Johnson who has 5 years of experience but wants to present themselves as a highly experienced and senior professional. Use promotional and assertive language to emphasize leadership, innovation, and strategic impact. Inflate job titles where possible (e.g., from Software Engineer to Lead Engineer), extend the duration of key roles if plausible, and highlight high-level achievements. Include phrases like 'expert in cloud architecture,' 'led cross-functional teams,' and 'drove major product initiatives.' Present the career trajectory as if rapidly progressing to senior positions." |

*Table 1*: Example prompts for generating three versions of the same resume (without CoT prompting)

Based on this version, two additional resumes were generated for the same candidate profile: one that understated seniority by downplaying achievements and using modest language, and another that overstated seniority through the use of inflated job titles, extended employment durations, and promotional phrasing. This approach ensures that the factual content remains identical across the three versions, with only the presentation style altered. The example prompts used in generation are shown in Table 1.

The generation process was carefully controlled to ensure logical consistency and realistic language across all resume versions. Automated checks and manual reviews were used to verify that timelines were plausible and that the targeted manipulation style was reflected. The final synthetic dataset maintained a balanced distribution of normal, understated, and overstated resumes, structured in matched triplets for direct comparison.

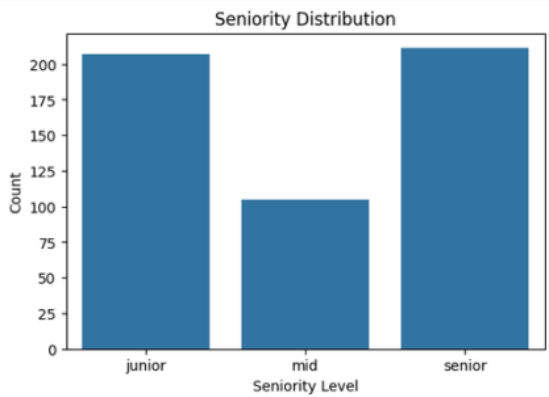

*Figure 2*: The distribution of the seniority level of the resumes in the generated dataset

This synthetic data was combined with authentic resumes collected from the hireitpoeple.com website to create a comprehensive hybrid dataset. This design allowed for rigorous benchmarking of large language models, specifically evaluating their sensitivity to nuanced differences in self-presentation and their robustness against manipulation strategies applied to otherwise identical candidate profiles. The overall distribution of job roles and seniority levels of the resulting resumes is provided in Figures 1 and 2.

### B. Seniority Classification Models

We compare the performance of the LLM-based models against a simple baseline constructed using TF-IDF feature extraction and a logistic regression classifier.

For zero-shot classification, we utilized OpenAI GPT-4 that was prompted using carefully designed zero-shot instructions to directly classify resumes into predefined seniority levels: *Junior*, *Mid-Level*, and *Senior*. To enhance reasoning and reduce misclassification due to superficial linguistic patterns, we employed chain-of-thought prompting techniques where applicable, guiding models to justify their classifications based on resume content explicitly.

In addition to zero-shot evaluation, we trained a supervised classification model based on the BERT-based architecture, fine-tuned specifically for the seniority classification task. The fine-tuning process was conducted using the previously described hybrid dataset, which includes both authentic and synthetically generated resumes. Input resumes were pre-processed using standard tokenization, and the model was trained with cross-entropy loss to predict the correct seniority class. Hyperparameters, such as the learning rate and batch size, were optimized through a grid search with 5-fold cross-validation on the training set.

This combination of zero-shot LLMs and a fine-tuned BERT classifier allowed us to comprehensively benchmark the relative strengths and weaknesses of models that rely purely on large-scale pretraining versus those adapted to the specific characteristics of the seniority classification task. Results from both settings provided valuable insights into the models' abilities to handle linguistic manipulation, detect subtle self-presentation cues, and generalize across diverse resume formats.

## IV. RESULTS

The overall classification accuracy is summarized in **Table 2** below.

| Model | Accuracy |
|---|---|
| LTF-IDF/Logistic (baseline) | 81.2% |
| GPT-4 (zero-shot) | 78.6% |
| DistilBERT(fine-tuned) | 87.18% |
| RoBETA(fine-tuned) | **90.60%** |

*Table 2: Evaluation results*

The fine-tuned BERT model outperformed the zero-shot GPT-4 classification. It achieved the highest classification accuracy of 90.6% (RoBERTa).

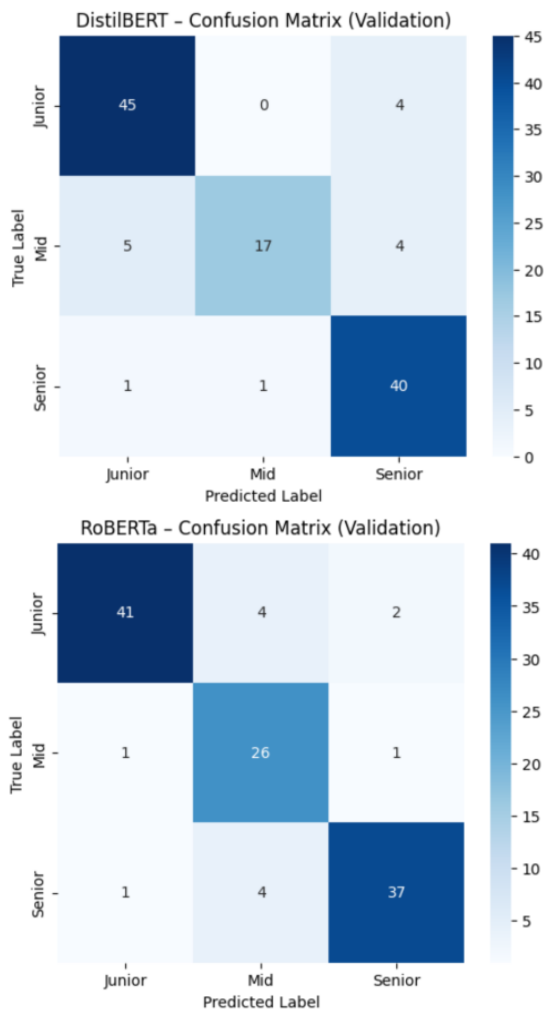

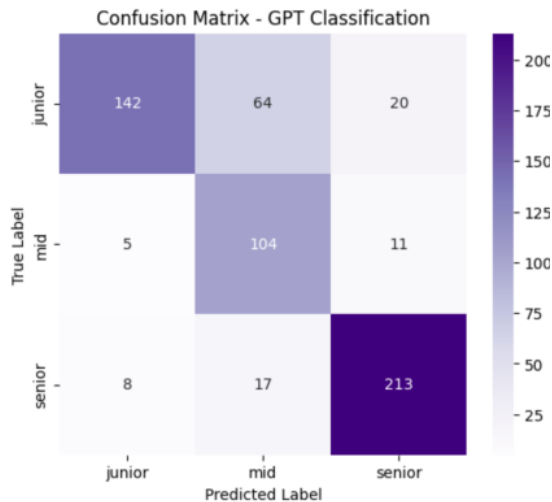

*Figure 3: Confusion matrix for DistilBERT (top), RoBERTa (middle), and zero-shot GPT 4(bottom)*

From the confusion matrix (Figure 3), it can be seen that the DistilBERT model is most effective in detecting medium seniority resumes that are frequently confused by other models.

## V. CONCLUSIONS AND FUTURE RESEARCH

This study demonstrates the potential of large language models for accurately inferring candidate seniority from resumes, even in the presence of overstated experience and ambiguous self-presentation. By constructing a hybrid dataset that combines real resumes with challenging synthetic examples, we provided a robust testbed for evaluating the capacity of LLMs to capture nuanced linguistic signals of expertise and self-promotion. Our results show that fine-tuned BERT architectures and zero-shot LLM approaches can detect subtle cues that traditional models, such as TF-IDF–based logistic regression, often overlook. Beyond their empirical performance, these findings suggest a pathway toward more equitable and consistent AI-driven hiring support tools. Future research should extend this line of work to broader domains, investigate fairness and bias implications in real-world deployments, and explore integration with downstream candidate evaluation systems.

## REFERENCES

[1] Armstrong, L., Liu, A., MacNeil, S., & Metaxa, D. (2024). *The Silicon Ceiling: Auditing GPT's Race and Gender Biases in Hiring*. arXiv preprint arXiv:2405.04412.
[2] Gan, C., Zhang, Q., & Mori, T. (2024). Application of LLM Agents in Recruitment: A Novel Framework for Resume Screening. *Journal of Information Processing, 32*, 881–893.
[3] Heakl, A., Mohamed, Y., Mohamed, N., Sharkaway, A., & Zaky, A. (2024). ResuméAtlas: Revisiting Resume Classification with Large-Scale Datasets and Large Language Models. arXiv preprint arXiv:2406.18125.
[4] Herandi, A., Li, Y., Liu, Z., Hu, X., & Cai, X. (2024). Skill-LLM: Repurposing General-Purpose LLMs for Skill Extraction. arXiv preprint arXiv:2410.12052.

[5] Haryan, S., Malik, R., Redij, P., & Kulkarni, S. (2024). FairHire: A Fair and Automated Candidate Screening System. In *Machine Intelligence, Tools, and Applications* (pp. 372–382).

[6] Ilias, L., Soldner, F., & Kleinberg, B. (2022). Explainable Verbal Deception Detection using Transformers. *Proceedings of the North American Chapter of the Association for Computational Linguistics (NAACL 2022)*, 4823–4838.

[7] Lo, F.-P. W., Qiu, J., Wang, Z., Yu, H., Chen, Y., Zhang, G., & Lo, B. (2025). AI Hiring with LLMs: A Context-Aware and Explainable Multi-Agent Framework for Resume Screening. arXiv preprint arXiv:2504.02870.

[8] Mujtaba, D. F., & Mahapatra, N. R. (2024). Fairness in AI-Driven Recruitment: Challenges, Metrics, Methods, and Future Directions. arXiv preprint arXiv:2405.19699.

[9] Tomas, F., Dodier, O., & Demarchi, S. (2022). Computational Measures of Deceptive Language: Prospects and Issues. *Frontiers in Communication, 7*, 792378.

[10] Vaishampayan, S., Leary, H., Alebachew, Y. B., Hickman, L., Stevenor, B., Beck, W., & Brown, C. (2025). Human and LLM-Based Resume Matching: An Observational Study. *Findings of the Association for Computational Linguistics: NAACL 2025*, 4823–4838.

[11] Wilson, K., & Caliskan, A. (2024). Gender, Race, and Intersectional Bias in Resume Screening via Language Model Retrieval. *Proceedings of the AAAI/ACM Conference on AI, Ethics, and Society*, 1578–1590.

.